\documentclass{article}

\usepackage{microtype}
\usepackage{graphicx}
\usepackage{subfigure}
\usepackage{booktabs} 

\usepackage{url}
\usepackage{graphicx}
\usepackage{natbib}
\usepackage{color}
\usepackage{subfigure}
\usepackage{nicefrac}
\usepackage{algorithmic}
\usepackage{algorithm}
\usepackage{enumitem}
\usepackage{amsmath}
\usepackage{amsthm}
\usepackage{amssymb}
\usepackage{bm}
\usepackage{multirow}

\newcommand{\NM}[2]{ \left\| #1 \right\|_{#2} }

\usepackage[accepted]{icml2020}

\usepackage{amsmath,amsfonts,bm}

\def\eqref#1{equation~\ref{#1}}

\def\1{\bm{1}}

\DeclareMathAlphabet{\mathsfit}{\encodingdefault}{\sfdefault}{m}{sl}
\SetMathAlphabet{\mathsfit}{bold}{\encodingdefault}{\sfdefault}{bx}{n}

\newtheorem{proposition}{Proposition}

\newtheorem{assumption}{Assumption}
\newtheorem{theorem}{Theorem}

\usepackage{array}
\newcolumntype{L}[1]{>{\raggedright\let\newline\\\arraybackslash\hspace{0pt}}m{#1}}
\newcolumntype{C}[1]{>{\centering\let\newline  \\\arraybackslash\hspace{0pt}}m{#1}}
\newcolumntype{R}[1]{>{\raggedleft\let\newline \\\arraybackslash\hspace{0pt}}m{#1}}

\icmltitlerunning{Searching to Exploit Memorization Effect in Learning with Noisy Labels}

\begin{document}

\twocolumn[
\icmltitle{Searching to Exploit Memorization Effect in Learning with Noisy Labels}




\begin{icmlauthorlist}
\icmlauthor{Quanming Yao}{4P}
\icmlauthor{Hansi Yang}{TU}
\icmlauthor{Bo Han}{BU,RK}
\icmlauthor{Gang Niu}{RK}
\icmlauthor{James T. Kwok}{UST}
\end{icmlauthorlist}

\icmlaffiliation{4P}{4Paradigm Inc (Hong Kong)}
\icmlaffiliation{TU}{Department of Electrical Engineering, Tshinghua University}
\icmlaffiliation{BU}{Department of Computer Science, Hong Kong Baptist University}
\icmlaffiliation{RK}{RIKEN Center for Advanced Intelligence Project}
\icmlaffiliation{UST}{Department of Computer Science and Engineering, Hong Kong University of Science and Technology}

\icmlcorrespondingauthor{Quanming Yao}{yaoquanming@4paradigm.com}


\vskip 0.3in
]



\printAffiliationsAndNotice{}  

\begin{abstract}
Sample selection approaches are popular in robust learning from noisy labels.  
However, 
how to properly control the selection process so that deep networks can benefit from the memorization effect is a hard problem.  
In this paper, 
motivated by the success of automated machine learning (AutoML),
we model this issue as a function approximation problem. 
Specifically,
we design a domain-specific search space based on general patterns of the memorization effect
and propose a novel Newton algorithm to solve the bi-level optimization problem efficiently.  
We further provide theoretical analysis of the algorithm,
which ensures a good approximation to critical points.
Experiments are performed on benchmark data sets.
Results demonstrate that the proposed method is 
much better than the state-of-the-art noisy-label-learning approaches, and also much more efficient than existing AutoML algorithms.
\end{abstract}

\section{Introduction}
\label{sec:intro}

Deep networks have enjoyed huge empirical success 
in a wide variety of tasks, such as
image processing, speech recognition,
language modeling and recommender systems \citep{goodfellow2016deep}.
However,
this highly counts on 
the availability of large amounts of quality data,
which may not be feasible
in practice.
Instead,
many large data sets 
are collected from 
crowdsourcing platforms 
or crawled from the internet,
and 
the obtained labels
are noisy 
\citep{Patrini2017}.
As deep networks
have large learning capacities,
they will eventually overfit the noisy labels,
leading to poor generalization performance
\citep{zhang2016understanding,arpit2017closer,jiang2017mentornet}.

To reduce the negative effects of noisy labels, 
a number of methods have been 
recently 
proposed 
\citep{sukhbaatar2014training,reed2014training,Patrini2017,ghosh2017robust,malach2017decoupling,liu2015classification,cheng2017learning}. 
They can be grouped into three main categories.
The first one is based 
on estimating the label transition matrix,
which captures how correct labels are flipped to the wrong ones
\citep{sukhbaatar2014training,reed2014training,Patrini2017,ghosh2017robust}.
However,
this can be fragile to heavy noise and is unable to handle a large number of labels
\citep{han2018co}. 
The second type is based on regularization \citep{miyato2017virtual,laine2016temporal,tarvainen2017mean}. 
However, 
since deep networks are usually over-parameterized,
they can still 
completely memorize the noisy data given sufficient training time \cite{zhang2016understanding}. 

The third approach, which is the focus in this paper, is based on 
selecting  (or weighting)
possibly clean 
samples  in each iteration
for training
\citep{jiang2017mentornet,han2018co,yu2019does,wang2019co}. 
Intuitively, by making the training data less noisy, better performance can be obtained.  
Representative methods include the MentorNet
\citep{jiang2017mentornet} 
and Co-teaching \citep{han2018co,yu2019does}.
Specifically, MentorNet uses
an additional network
to select clean samples for training of a StudentNet. 
Co-teaching improves MentorNet by
simultaneously maintaining two networks with identical architectures during 
training, and
each network is updated using the 
small-loss samples
from the other network.

In sample selection, a core issue 
is how many small-loss samples are to be
selected in each iteration.  While discarding a lot of samples can avoid training with noisy labels, 
 dropping
 too many can 
 be overly conservative 
 and
 lead to lower accuracy 
 \citep{han2018co}. 
Co-teaching uses the observation that
deep networks usually learn easy patterns before 
overfitting the noisy samples \citep{zhang2016understanding,arpit2017closer}. 
This \textit{memorization effect} 
has been widely seen in various
deep networks
\cite{Patrini2017,ghosh2017robust,han2018co}.
Hence, during the early stage of training,  Co-teaching drops very few samples as
the network will not memorize the noisy data.
As training proceeds, the network starts to memorize the noisy data. This is 
avoided in Co-teaching by gradually dropping more samples according to a pre-defined schedule.
Empirically, this signiificantly improves the network's generalization performance on noisy labels
\citep{jiang2017mentornet,han2018co}.
However, it is unclear if its manally-designed schedule 
is ``optimal". 
Moreover, the schedule is not data-dependent, but is the same for all data sets.
Manually 
finding a good 
schedule 
for each and every data set 
is clearly very time-consuming and infeasible.

Motivated by the recent success of automated machine learning
(AutoML) \citep{automl_book,quanming2018auto}, 
in this paper
we propose to exploit the memorization effect
automatically using AutoML.  We first formulate the learning of schedule as a
bi-level optimization problem, similar to that in  neural architecture search (NAS)
\citep{zoph2017neural}.  A search space for the schedule is designed based on the
learning curve behaviors shared by deep networks.  This space is expressive, and
yet compact  with only a small number of hyperparameters. However,
computing the gradient is difficult as sample selection is a discrete operator.
To 
avoid this problem and
perform efficient search,
we propose to use stochastic relaxation 
\cite{geman1984stochastic}
together with Newton's method to capture
information from both the model and optimization objective.
Convergence analysis is provided, and 
extensive experiments 
are performed 
on benchmark datasets.
Empirically, the proposed method 
outperforms
state-of-the-art methods, and can select a higher proportion of clean samples than
other sample selection methods.
Ablation studies show that the chosen search space is appropriate, and
the proposed search algorithm is faster than 
popular 
AutoML search algorithms in this context.

\textbf{Notation.}
In the sequel,
scalars are in lowercase letters,
vectors are in lowercase boldface letters,
and matrices  are in uppercase boldface letters.
The gradient of a function $\mathcal{J}$
is denoted 
$\nabla \mathcal{J}$, and
$\NM{\cdot}{}$ denotes the $\ell_2$-norm of a vector.

\section{Related work}
\label{work}

\begin{figure*}[ht]
	\centering
	\subfigure[Impact of $R(t)$. \label{fig:1a}]
	{\includegraphics[width=0.32\textwidth]{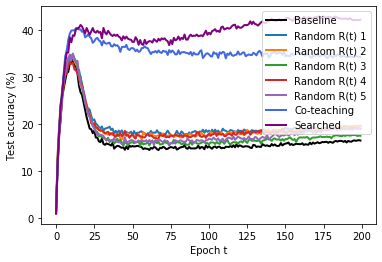}
	\label{fig:mem:rt}}
	\subfigure[Different data sets (training accuracy). \label{fig:1b}]
	{\includegraphics[width=0.32\textwidth]{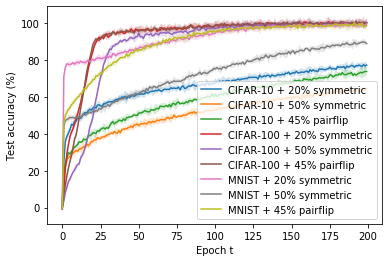}
		\label{fig:mem:tra}}
	\subfigure[Different data sets (testing accuracy). \label{fig:1c}]
	{\includegraphics[width=0.32\textwidth]{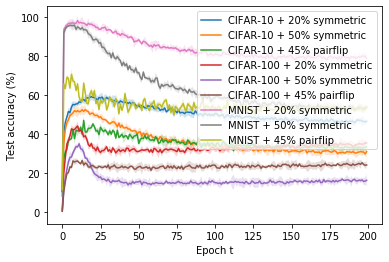}
		\label{fig:mem:test}}
	
	\subfigure[Different architectures. \label{fig:1d}]
	{\includegraphics[width=0.32\textwidth]{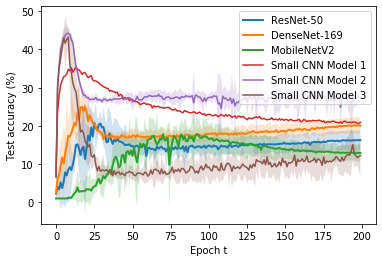}
	\label{fig:mem:model}}
	\subfigure[Different optimizers. \label{fig:1e}]
	{\includegraphics[width=0.32\textwidth]{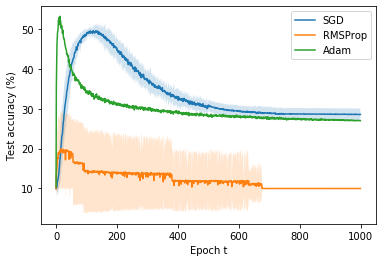}
		\label{fig:mem:opt}}
	\subfigure[Different optimizer settings. \label{fig:1f}]
	{\includegraphics[width=0.32\textwidth]{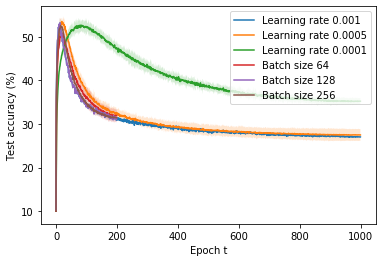}
		\label{fig:mem:rate}}
	
	\caption{Training and testing accuracies on 
		CIFAR-10, CIFAR-100, and
		MNIST
		using various architectures,
		optimizers,
		and optimizer settings. The 
		detailed setup is in Appendix~\ref{app:fig1}.}
	\label{fig:dif:mem}
\end{figure*}

\subsection{Automated Machine Learning (AutoML)}
\label{sec:rel:automl}

Recently,
AutoML
has 
shown to be very
useful in the design of machine learning models
\citep{automl_book,quanming2018auto}.
Two of its important ingredients 
are:
\begin{enumerate}[topsep=0pt,parsep=0pt,partopsep=0pt,leftmargin=11px]
\item \textit{Search space}, which needs to be specially designed for each AutoML problem
	\cite{baker2017designing,liu2018darts,zhang2020efficient}.   
	It
	should be general (so as to 
	cover existing models), 
	yet not too general (otherwise searching in this space will be expensive).
	
	\item \textit{Search algorithms}: Two types are popularly used.  The first
	includes
	derivative-free optimization methods, such as
	reinforcement learning \citep{zoph2017neural,baker2017designing},
	genetic programming \citep{xie2017genetic},
	and Bayesian optimization \citep{bergstra2011algorithms,snoek2012practical}.
	The second type is gradient-based, and
	updates the  parameters and hyperparameters
	in an alternating manner. 
	On NAS problems,
	gradient-based methods are 
	usually
	more efficient than
	derivative-free methods
	\citep{liu2018darts,akimoto2019adaptive,yao2020efficient}.
\end{enumerate}

\subsection{Learning from Noisy Labels}
\label{sec:rel:noisy}

The state-of-the-arts usually combat noisy labels by
sample selection
\citep{jiang2017mentornet,han2018co,malach2017decoupling,yu2019does,wang2019co},
which only uses the ``clean" samples 
(with relatively small losses)
from each mini-batch for training.
The general procedure is shown in Algorithm~\ref{alg:gen}. 
Let
$f$ 
be the classifier to be learned.
At the $t$th iteration,
a subset $\bar{\mathcal{D}}_f$ of small-loss samples
are selected from the mini-batch $\bar{\mathcal{D}}$ (step~3).
These ``clean'' samples are then used to update the
network parameters in step~4.

\begin{algorithm}[ht]
\caption{General procedure on using sample selection to combat noisy labels.}
	\begin{algorithmic}[1]
		\FOR{$t = 0, \dots, T-1$}
		\STATE draw a mini-batch $\bar{\mathcal{D}}$ from $\mathcal{D}$;
		\STATE select $R(t)$ small-loss samples $\bar{\mathcal{D}}_f$ from $\bar{\mathcal{D}}$ based on network's predictions;
		\STATE update network parameter using 
		$\bar{\mathcal{D}}_f$;
		\ENDFOR
	\end{algorithmic}
	\label{alg:gen}
\end{algorithm}

\section{Proposed Method}
\label{headings}

\subsection{Motivation}
\label{sec:motivate}

In step~3 of Algorithm~\ref{alg:gen},
$R(\cdot)$ 
controls how many samples are 
selected into $\bar{\mathcal{D}}_f$.
As can be seen from
Figure~\ref{fig:mem:rt},
its setting is often critical to the performance,
and random $R(t)$ schedules have only marginal improvements over
directly training on the whole noisy data set
(denoted ``Baseline" in the figure)
\citep{han2018co,ren2018learning}.
Moreover, while having a large
$R(\cdot)$  can avoid training with noisy labels, 
dropping too many samples can lead to lower accuracy, as demonstrated in Table~8 of \citep{han2018co}.

Based on the memorization effect in deep networks \cite{zhang2016understanding},
Co-teaching \citep{han2018co} (and its variant
Co-teaching+ \citep{yu2019does}) 
designed the  following
schedule:
\begin{equation}
\label{eq:han}
R(t) = 1 - \tau \cdot \min( (t/t_k)^c, 1 ),
\end{equation}
where $\tau$, $c$ and $t_k$ are some hyperparameters.
As
can be seen from Figure~\ref{fig:mem:rt},
it can significantly improve the performance  over random schedules.

While $R(\cdot)$ is critical and that
it is important to exploit the memorization effect,
it is unclear if 
the schedule in 
(\ref{eq:han}) is ``optimal". Moreover, the same 
schedule is used by 
Co-teaching 
on all the  data sets. This is expected to be suboptimal, but 
it is hard to find $R(\cdot)$  for each and every data set manually.
This
motivates us to formulate  the design of $R(\cdot)$ as an AutoML problem
that 
searches for a good
$R(\cdot)$
automatically 
(Section~\ref{sec:formulate}). 
The two
important ingredients of AutoML, namely, search space and
search algorithm, will then be described in Sections~\ref{sec:space} and
\ref{sec:alg}, respectively.

\subsection{Formulation as an AutoML Problem}
\label{sec:formulate}

Let the noisy training (resp. clean
validation) data set be $\mathcal{D}_{\text{tr}}$ (resp.
$\mathcal{D}_{\text{val}}$), the
training (resp. validation) loss be $\mathcal{L}_{\text{tr}}$ (resp. $\mathcal{L}_{\text{val}}$),
and $f$ be a neural network with model parameter $w$.
We formulate the design of $R(\cdot)$ 
in Algorithm~\ref{alg:gen}
as the following AutoML problem:
\begin{align}
R^*  
& = \underset{ R(\cdot) \in \mathcal{F} }{ \arg\min }
\mathcal{L}_{\text{val}} (f(\bm{w}^*; R), \mathcal{D}_{\text{val}}),
\label{eq:automl:high}
\\
\text{s.t.\;} 
\bm{w}^* 
& = 
\underset{ \bm{w} }{ \arg\min }
\mathcal{L}_{\text{tr}}(f(\bm{w}; R), \mathcal{D}_{\text{tr}}).
\label{eq:automl:low}
\end{align}
where
$\mathcal{F}$  is
the search space 
of $R(\cdot)$.

Similar to the AutoML problems of
auto-sklearn \cite{feurer2015efficient} and NAS \cite{zoph2017neural,liu2018darts,yao2020efficient}, 
this is also a bi-level optimization problem \cite{colson2007overview}. 
At the outer level (subproblem~(\ref{eq:automl:high})), 
a good $R(\cdot)$ is searched based on the validation set. 
At the lower level (subproblem~(\ref{eq:automl:low})), 
we find the model parameters 
using the training set.

\subsection{Designing the Search Space
$\mathcal{F}$} 
\label{sec:space}

In Section~\ref{sec:obv}, we first discuss some
observations from  the learning curves of deep networks.
These are then used in the design of an appropriate search space for
$R(\cdot)$
in
Section~\ref{sec:prior}.

\subsubsection{Observations from Learning Curves}
\label{sec:obv}

Figures~\ref{fig:mem:tra}-\ref{fig:mem:rate} show 
the training and validation set accuracies obtained on 
the MNIST, CIFAR-10, CIFAR-100
data sets,
which are
corrupted with 
different types and levels
of label noise 
(symmetric flipping 20\%, symmetric flipping 50\%, and pair flipping 45\%), 
using a number of architectures
(ResNet \cite{he2016deep}, DenseNet \cite{huang2017densely} and small CNN models in
\cite{yu2019does}),
optimizers
(SGD \cite{bottou2010large}, 
Adam \cite{kingma2014adam} and RMSProp \cite{hinton2012neural})
and optimizer settings (learning rate and batch size). 

As can be seen,
the training accuracy always increases as training progresses
		(Figure~\ref{fig:mem:tra}),
while the testing 
accuracy first increases 
and then slowly drops
due to over-fitting
(Figure~\ref{fig:mem:test}).
Note that this pattern is independent of the network architecture (Figure~\ref{fig:mem:model}), choice of optimizer (Figure~\ref{fig:mem:opt}), and hyperparameter (Figure~\ref{fig:mem:rate}).

Recall that 
deep networks
usually 
learn easy patterns  
first 
before
memorizing and overfitting
the noisy samples \citep{arpit2017closer}. 
From (\ref{eq:han}) and Figure~\ref{fig:dif:mem},
we have the following observations on $R(t)$:
\begin{itemize}[leftmargin=*]
\item 
During the initial phase when the learning curve rises,
the deep network is plastic and 
can learn easy patterns   from the data.
In this phase, one can allow a larger $R(t)$ as there is little risk of
memorization.
Hence,
at time $t=0$, 
we can set $R(0) = 1$ and the entire noisy data set is used.

\item 
As training proceeds and the learning curve has peaked, 
the network starts to
memorize and overfit
the noisy samples.
Hence, $R(t)$ should then decrease.
As can be seen from 
Figure~\ref{fig:mem:rt}, this can significantly improve the 
network's generalization performance 
on noisy labels.

\item 
Finally, 
as the network gets less plastic
and in case $R(t)$ drops too much at the beginning,  
it may be useful to allow $R(t)$ 
to slowly increase so as to enable learning.
\end{itemize}


The above motivates us to impose the following prior knowledge on 
the search space $\mathcal{F}$ of $R(\cdot)$.
An example 
$R(\cdot)$
is shown in Figure~\ref{fig:expfi}.

\begin{assumption}[A Prior on $\mathcal{F}$] \label{ass:prior}
The shape of $R(\cdot)$ should be opposite to that of the learning curve. 
Besides, 
as in \cite{han2018co,yu2019does},
it is natural to have $R(t) \in [0,1]$ and
$R(0) = 1$.  
\end{assumption}

\subsubsection{Imposing Prior Knowledge}
\label{sec:prior}

\begin{figure}[ht]
\centering
\includegraphics[width=0.30\textwidth]{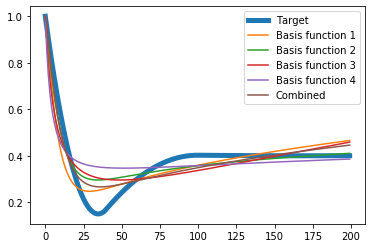}

\caption{Plots of the basis functions in Table~\ref{tab:expfi}.
An example $R(\cdot)$ to be learned is shown in blue.}
\label{fig:expfi}
\end{figure}

To allow efficient search,
the search space 
has to be
small but not too small. To achieve this, we impose
the prior knowledge 
proposed in Section~\ref{sec:obv}
on $\mathcal{F}$.
Specifically, we
use $k$
basis functions 
($f_i$'s)
whose shapes follow Assumption~\ref{ass:prior}
(shown in Table~\ref{tab:expfi} and Figure~\ref{fig:expfi}).
The exact choice of these
basis functions is not important.
The search space for $R(\cdot)$ 
is then defined 
as:
\begin{align}
\!\!\!\!
\mathcal{F} 
\equiv
\left\{
R(t) =
\sum_{i = 1}^k \alpha_i f_i(t; \beta_i)
:
\sum_{i} \alpha_i = 1, \alpha_i \ge 0
\right\},
\!\!
\label{eq:rtalpha}
\end{align}
where 
$\beta_i$ is the hyperparameter associated with basis function $f_i$. 
In the experiments, we set all $\beta_i$'s to be in the range $[0,1]$. 
Let $\bm{\alpha} = \{ \alpha_i \}$,
$\bm{\beta} = \{ \beta_i \}$
and $\bm{x} \equiv \{ \bm{\alpha}, \bm{\beta} \}$.
The search algorithm to be introduced 
will then only need
to search for 
a small set
of hyperparameters 
$\bm{x}$.

\begin{table}[ht]
\renewcommand{\arraystretch}{1.40}
\centering
\caption{The four basis functions used to define the search space in the experiments. Here, $a_i$'s are the hyperparameters.}
\begin{tabular}{C{20px} | C{120px}}
	\hline
	$f_1$ & $e^{-a_2 t^{a_1}} + a_3(\frac{t}{T})^{a_4}$                             \\ \hline
	$f_2$ & $e^{-a_2 t^{a_1}} + a_3 \frac{\log(1+t^{a_4})}{\log(1+{T}^{a_4})}$      \\ \hline
	$f_3$ & $\frac{1}{(1+a_2 t)^{a_1}}+a_3(\frac{t}{T})^{a_4}$                      \\ \hline
	$f_4$ & $\frac{1}{(1+a_2t)^{a_1}} + a_3 \frac{\log(1+t^{a_4})}{\log(1+{T}^{a_4})}$ \\ \hline
\end{tabular}
\label{tab:expfi}
\end{table}

With $\mathcal{F}$ in (\ref{eq:rtalpha}),
the outer problem 
in (\ref{eq:automl:high})
becomes
\begin{equation} \label{eq:new}
\left\lbrace 
\bm{\alpha}^*, \bm{\beta}^*
\right\rbrace 
= 
\underset{ R(\cdot) \in \mathcal{F} }{ \arg\min } \;
\mathcal{L}_{\text{val}} (f(\bm{w}^*; R), \mathcal{D}_{\text{val}}),
\end{equation} 
and the optimal $R^*$ in (\ref{eq:automl:high}) is $\sum_{i = 1}^k \alpha_i^* f_i(t; \beta_i^*)$.

\subsubsection{Discussion}

As will be shown in Section~\ref{sec:aba:space},
the search space 
used in 
Co-teaching and Co-teaching+
is not large enough to ensure good performance.
Besides,
the design of $R(t)$ can be considered as a 
function learning problem, and
general function approximators
(such as radial basis function networks and multilayer perceptrons) can also be
used.
However, as
will be demonstrated
in Section~\ref{sec:aba:space},
the resultant search space is too large for 
efficient search,
while the prior on $\mathcal{F}$ in (\ref{eq:rtalpha}) can 
provide satisfactory performance.
Note that the proposed search space can well approximate
the space in Co-teaching (details are in Appendix~\ref{app:appco}).

\subsection{Search Algorithm Based on Relaxation}
\label{sec:alg}

Gradient-based methods 
\cite{bengio2000gradient,liu2018darts,yao2020efficient}
have been popularly used  in NAS and hyperparameter optimization.
Usually, the gradient 
w.r.t. 
hyperparameter
$\bm{x}$
is computed
via the chain rule as:
$\nabla_{\bm{x}} \mathcal{L}_{\text{val}}
= \nabla_{ \bm{w}^* } \mathcal{L}_{\text{val}} \cdot \nabla_{\bm{x}} \bm{w}^*$.
However, $\nabla_{\bm{x}} \bm{w}^*$ is hard to obtain here,
as the hyperparameters in $R(\cdot)$ control the selection of samples in each
mini-batch, a discrete operation.

\subsubsection{Stochastic Relaxation
with 
Newton's Method 
}
\label{ssec:natu}

To avoid a direct computation of the gradient w.r.t $\bm{x}$,
we propose to transform problem (\ref{eq:automl:high}) with stochastic relaxation
\cite{geman1984stochastic}.
This has also been recently explored in AutoML \cite{baker2017designing,pham2018efficient,akimoto2019adaptive}.
Specifically,
instead
of (\ref{eq:automl:high}), we
consider the following optimization problem:
\begin{align}
\min_{\bm{\theta}} \mathcal{J}(\bm{\theta})
& \equiv \int_{ \bm{x} \in \mathcal{F} } 
\bar{f}(\bm{x})
p_{\bm{\theta}}( \bm{x} ) \;d \bm{x},
\label{eq:obj}
\end{align}
where 
$\bar{f}(\bm{x}) \equiv \mathcal{L}_{\text{val}}( f(\bm{w}^*; R(\bm{x})),
\mathcal{D}_{val}) $ in (\ref{eq:new}), and
$p_{\bm{\theta}}( \bm{x} )$ is a distribution 
(parametrized by $\bm{\theta}$) 
on the search space $\mathcal{F}$
in (\ref{eq:rtalpha}).
As $\alpha_i \ge 0$ and $\sum_{i} \alpha_i=1$,
we use the Dirichlet distribution
on $\bm{\alpha}$.
We use
the Beta distribution 
on $\bm{\beta}$, 
as each $\beta_i$ lies in a bounded interval.
Note that minimizing $\mathcal{J}(\bm{\theta})$ coincides with minimization of
(\ref{eq:automl:high}),
i.e., $\min_{\bm{\theta}} \mathcal{J}(\bm{\theta}) = \min_{\bm{x}} \bar{f}(\bm{x})$ \cite{akimoto2019adaptive}.

Let $\bar{\bm{p}}_{\bm{\theta}}(\bm{x}) \equiv \nabla \log p_{\bm{\theta}}(\bm{x})$. 
As $\mathcal{J}(\bm{\theta})$  is smooth, it
can be minimized by gradient descent, with 
\begin{align*}
\nabla \mathcal{J}(\bm{\theta})
= \int_{\bm{x} \in \mathcal{F} } 
\bar{f}(\bm{x})
\nabla p_{\bm{\theta}}(\bm{x}) d \bm{x}
= \mathbb{E}_{ p_{\bm{\theta}} }
\left[ \bar{f}(\bm{x}) \bar{\bm{p}}_{\bm{\theta}} (\bm{x}) \right].
\end{align*}
The expectation can be approximated 
by sampling $K$ $\bm{x}_i$'s from $p_{\bm{\theta}}(\cdot)$,
leading to
\begin{equation} \label{eq:grad2}
\nabla \mathcal{J}(\bm{\theta}) \simeq \frac{1}{K} \sum_{i = 1}^K \bar{f}(\bm{x}_i) 
\bar{\bm{p}}_{\bm{\theta}}(\bm{x}_i).
\end{equation} 
The update at the $m$th iteration is then
\begin{align}
\bm{\theta}^{m+1}=\bm{\theta}^m + \rho  \bm{H}^{-1}\nabla \mathcal{J}(\bm{\theta}^m),
\label{eq:rule}
\end{align}
where $\rho$ is the stepsize,  
$\bm{H} = \bm{I}$ for gradient descent
and
$\bm{H} =  \mathbb{E}_{ p_{\bm{\theta}^m} }
[ \bar{\bm{p}}_{\bm{\theta}}(\bm{x}) \bar{\bm{p}}_{\bm{\theta}}(\bm{x})^{\top}]$ (i.e., Fisher matrix)
for natural gradient descent.

In general, natural gradient considers the geometrical structure  of the underlying probability
manifold,
and is more efficient than simple gradient descent. However,  here, the manifold is
induced by a 
$p_{\bm{\theta}}$ that is artificially introduced for
stochastic relaxation. Subsequently,
the Fisher matrix 
is independent of the objective 
$\mathcal{J}$.
In this paper, we 
instead
propose to use the Newton's method  and set
$\bm{H} =  
\nabla^2 \mathcal{J}(\bm{\theta})$,
which explicitly 
takes 
$\mathcal{J}$ into account.
The following Proposition shows that the Hessian can be easily computed
(proof is in Appendix~\ref{app:proofs}),
and
clearly 
incorporates more information than the Fisher matrix.
Moreover, it can also be approximated with finite samples as in (\ref{eq:grad2}).
\begin{proposition}\label{prop:Hessian}
$\nabla^2 \mathcal{J}(\bm{\theta}) = \mathbb{E}_{p_{\bm{\theta}}} \left[ \bar{f} (\bm{x})\nabla^2 \log p_{\bm{\theta}}(\bm{x})  \right]  
+ \mathbb{E}_{p_{\bm{\theta}}}\left[ \bar{f} (\bm{x}) \bar{\bm{p}}_{\bm{\theta}}(\bm{x})  \bar{\bm{p}}_{\bm{\theta}}(\bm{x})^{\top} \right]$.
\end{proposition}

The whole procedure,
which will be called {\em Search to Exploit\/} (S2E), is shown in Algorithm~\ref{alg:newton}.

\begin{algorithm}[ht]
\caption{{\em Search to Exploit\/} (S2E) algorithm
for the minimization of the relaxed objective 
$\mathcal{J}$ 
in (\ref{eq:obj}).}
\begin{algorithmic}[1]
\STATE Initialize $\bm{\theta}^1=\bm{1}$ so that $p_{\bm{\theta}}(\bm{x})$ is uniform distribution. 
	\FOR{$m = 1, \dots, M$}
	\FOR{$k = 1, \dots, K$}
	\STATE draw hyperparameter $\bm{x}$ from distribution $p_{\bm{\theta}^m}(\bm{x})$; 
	\STATE 
using $\bm{x}$,
	run Algorithm~\ref{alg:gen} with $R(\cdot)$ in (\ref{eq:rtalpha});
	\ENDFOR
	\STATE use the $K$ samples in steps~3-6 to approximate $\nabla
	\mathcal{J}(\bm{\theta}^m)$ 
	in (\ref{eq:grad2}) and 
	$\nabla^2 \mathcal{J}(\bm{\theta}^m) $ in
	Proposition~\ref{prop:Hessian};
	\STATE update $\bm{\theta}^m$ by (\ref{eq:rule});
	\ENDFOR
\end{algorithmic}
\label{alg:newton}
\end{algorithm}

\subsubsection{Convergence Analysis}
\label{sec:convana}

When $K = \infty$ in (\ref{eq:grad2}),
classical analysis \cite{rockafellar1970convex}
ensures that Algorithm~\ref{alg:newton} converges at a
critical point of (\ref{eq:obj}).
When $\mathcal{J}$ is convex,
a super-linear convergence rate is also guaranteed.
However,
when $K \not= \infty$,
the approximation of $\nabla \mathcal{J}(\bm{\theta})$
in (\ref{eq:grad2}) and the analogous
approximation of $\nabla^2 \mathcal{J}(\bm{\theta})$
introduce errors into the gradient.
To make this explicit, we rewrite
(\ref{eq:rule}) as
\begin{align}
\bm{\theta}^{m + 1}
= \bm{\theta}^{m} 
\! - \! ( \bm{\Delta}^m )^{-1} 
( \nabla \mathcal{J}(\bm{\theta}^m) \! - \! \bm{e}^m ),
\label{eq:app1}
\end{align}
where $\bm{\Delta}^m$ and $\bm{e}^m$ are the approximated  
Hessian 
and gradient errors, respectively, at the $m$th iteration.

We make the following Assumption on $\mathcal{J}$,
which requires $\mathcal{J}$ to be smooth and bounded from below.
\begin{assumption} \label{ass:func}
(i) $\mathcal{J}$ is $L$-Lipschitz smooth,
i.e.,
$\NM{\nabla \mathcal{J}(\bm{x}) - \nabla \mathcal{J}(\bm{y}) }{}
\le L \NM{\bm{x} - \bm{y}}{}$ 
for some
positive 
$L$;
(ii) $\mathcal{J}$ is coercive,
i.e., $\inf_{\bm{\theta}} \mathcal{J}(\bm{\theta}) > - \infty$ 
and 
$\lim\nolimits_{ \NM{\bm{\theta}}{} \rightarrow \infty } \mathcal{J}(\bm{\theta}) = \infty$.
\end{assumption}

We make the following Assumption~\ref{ass:error} on (\ref{eq:app1}).
Note that $\bar{\varepsilon} = 0$ when $K \rightarrow \infty$.
However,
since $K \not= \infty$ in practice,
the errors in $\bm{\Delta}^m$ and $\bm{e}^m$ do not vanish,
i.e., $\lim\nolimits_{m \rightarrow \infty} \left[ \bm{\Delta}^m -\nabla^2 \mathcal{J}(\bm{\theta}^m) \right] \not= \bm{0}$
and $\lim\nolimits_{m \rightarrow \infty} \bm{e}^m \not= \bm{0}$,
Assumption~\ref{ass:error} is 
more relaxed than the typical vanishing error assumptions
used in classical analysis of
first-order optimization algorithms \cite{schmidt2011convergence,bolte2014proximal,yao2016efficient}.

\begin{assumption} \label{ass:error}
(i)
$\eta \le \sigma(\bm{\Delta}^m) \le L$, where $\sigma(\cdot)$ denotes
eigenvalues of the matrix argument,
and $\eta$ is a positive constant;
(ii) 
Gradient 
errors 
are bounded:
$\forall m$, $\NM{ \bm{e}^m }{} \! \le \! \bar{\varepsilon}$.
\end{assumption}

Using Assumptions~\ref{ass:func} and \ref{ass:error},
the following Proposition 
bounds the difference in objective values at two consecutive iterations.
Note that the RHS below may not be positive, and so 
$\mathcal{J}$ may not be non-increasing.

\begin{proposition} \label{pr:twoiter}
$\mathcal{J}( \bm{\theta}^m ) 
- 
\mathcal{J}( \bm{\theta}^{m + 1} )
\ge 
\frac{2 - L \eta}{2 \eta}
\NM{ \bm{\gamma}^m }{}^2
- \NM{ \bm{e}^m }{} \NM{ \bm{\gamma}^m }{}$,
where $\bm{\gamma}^m = \bm{\theta}^{m + 1} - \bm{\theta}^m$.
\end{proposition}

\begin{figure*}[ht]
	\centering
	
	{\includegraphics[width=0.95\textwidth]{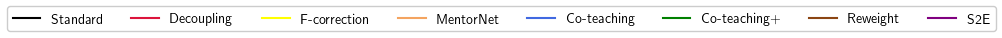}}
	
	{\includegraphics[width=0.32\textwidth]{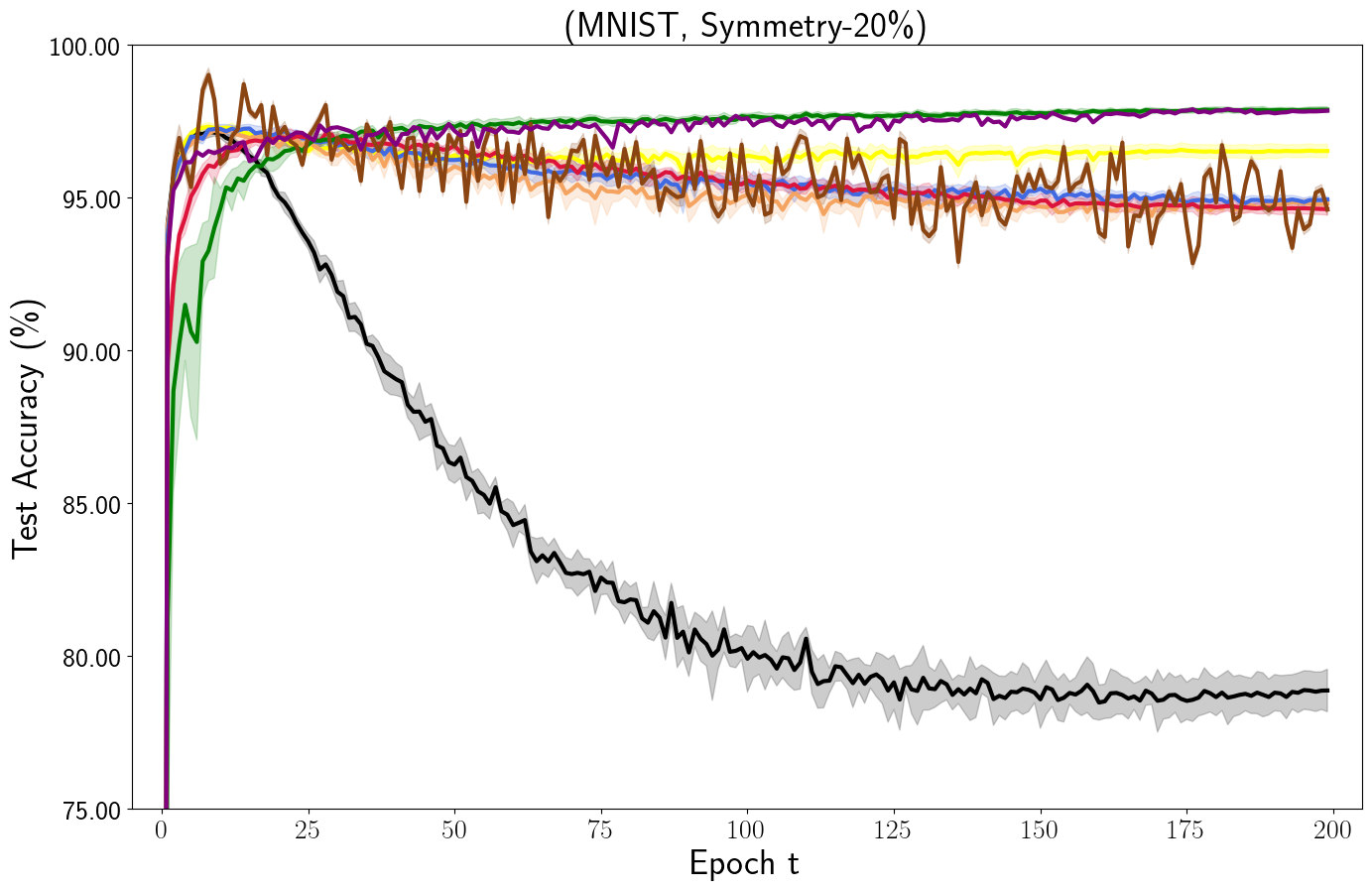}}
	{\includegraphics[width=0.32\textwidth]{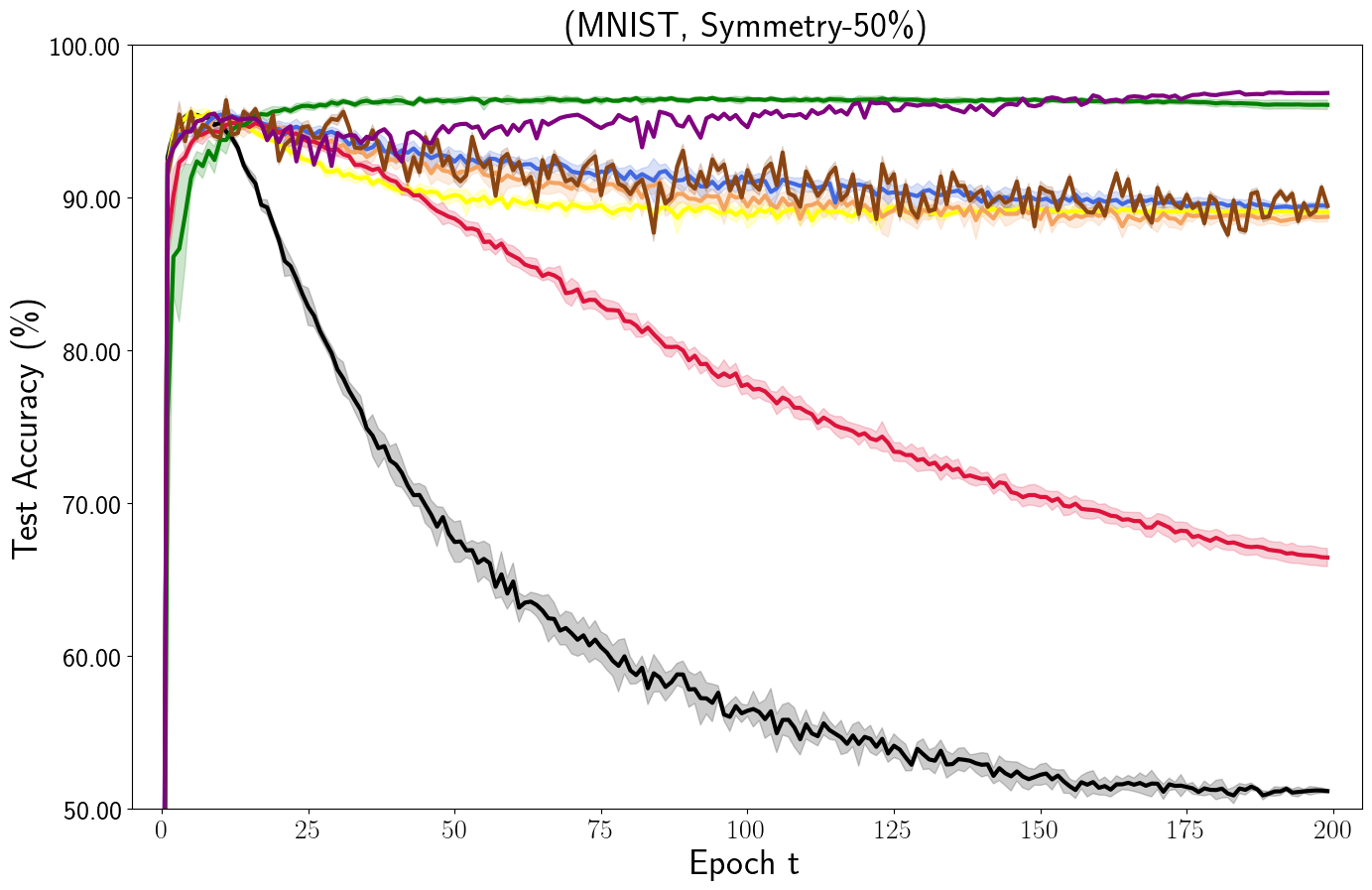}}
	{\includegraphics[width=0.32\textwidth]{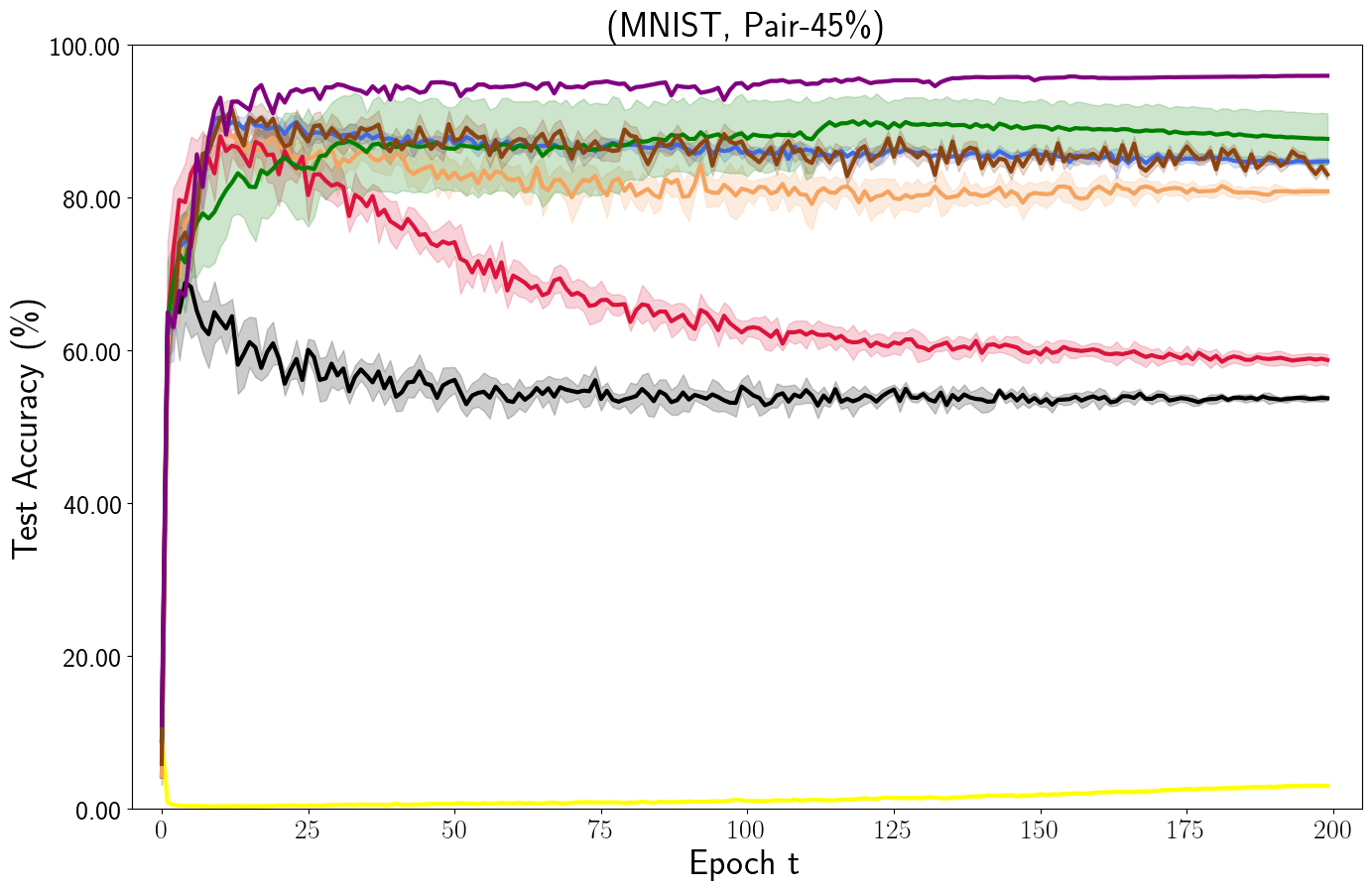}}

	{\includegraphics[width=0.32\textwidth]{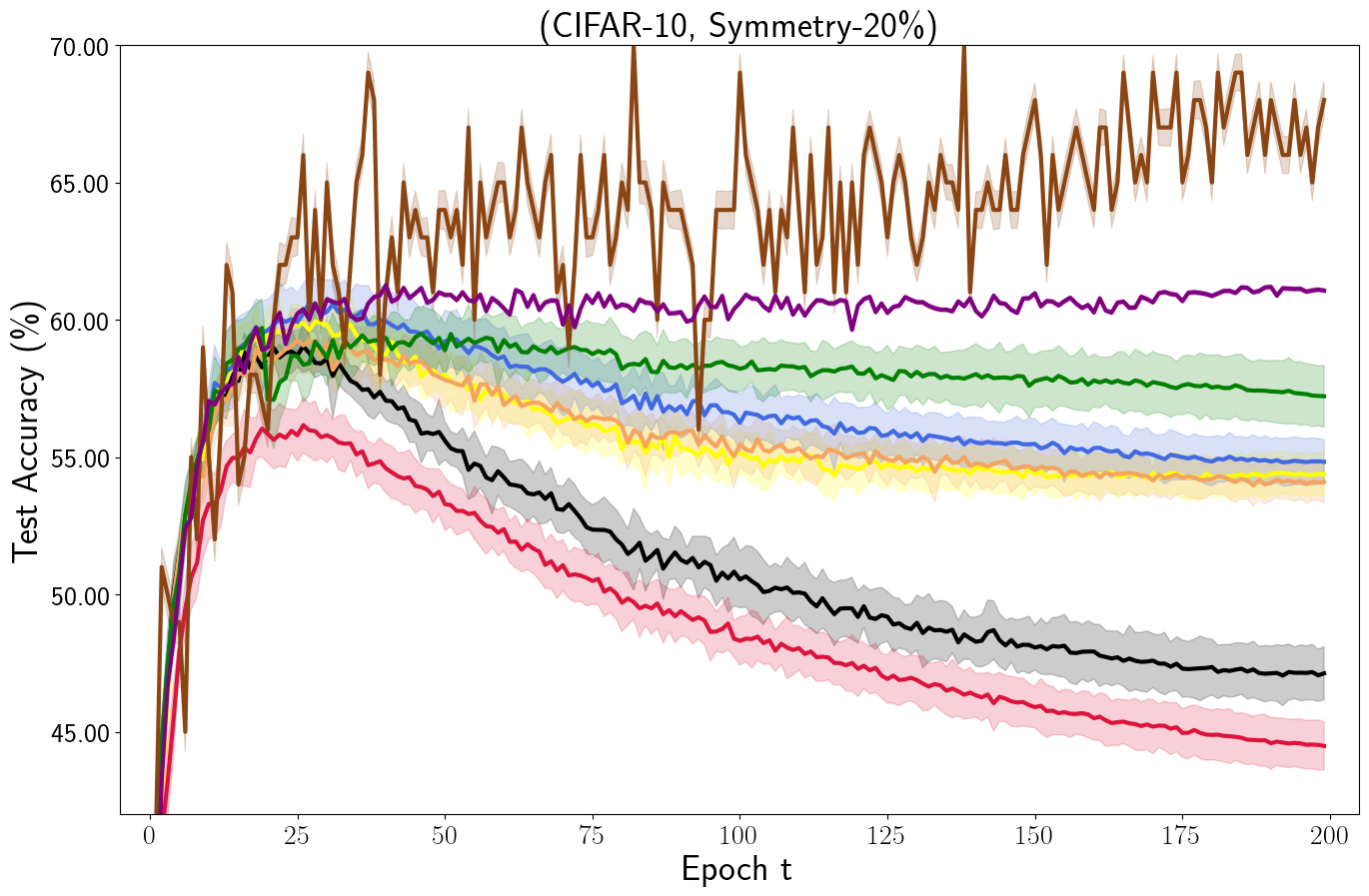}}
	{\includegraphics[width=0.32\textwidth]{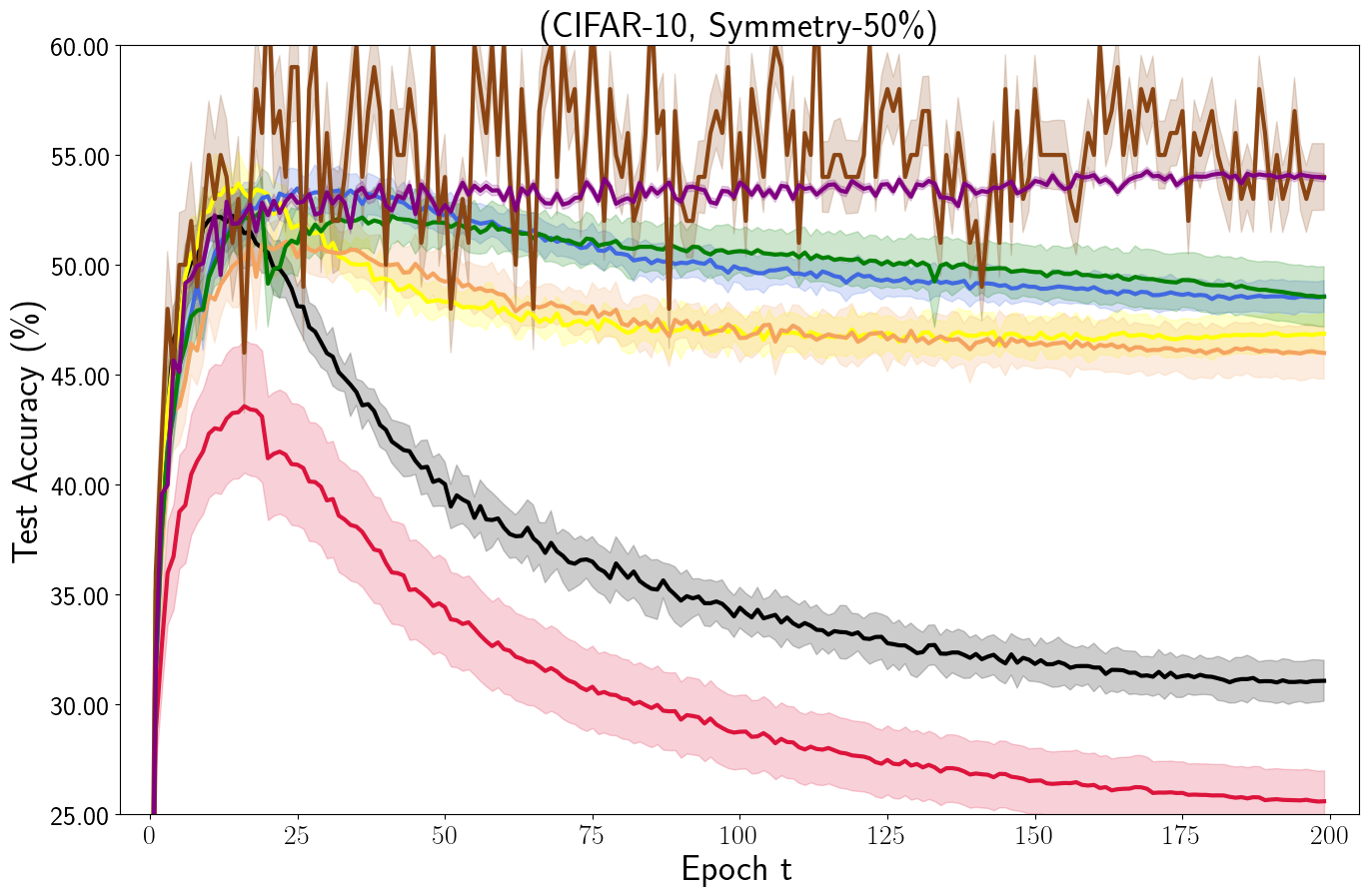}}
	{\includegraphics[width=0.32\textwidth]{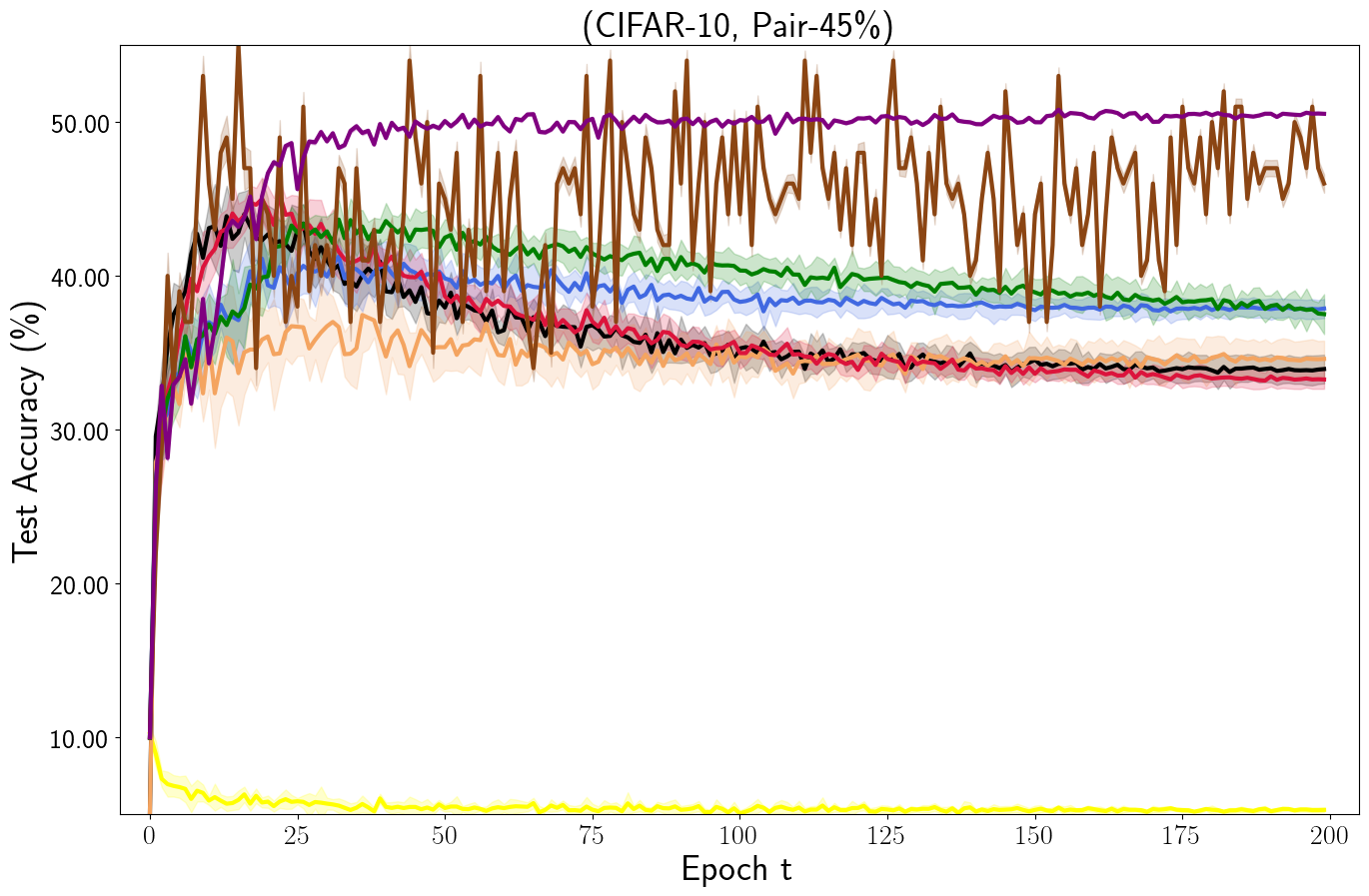}}

	\subfigure[symmetry flipping (20\%).]
	{\includegraphics[width=0.32\textwidth]{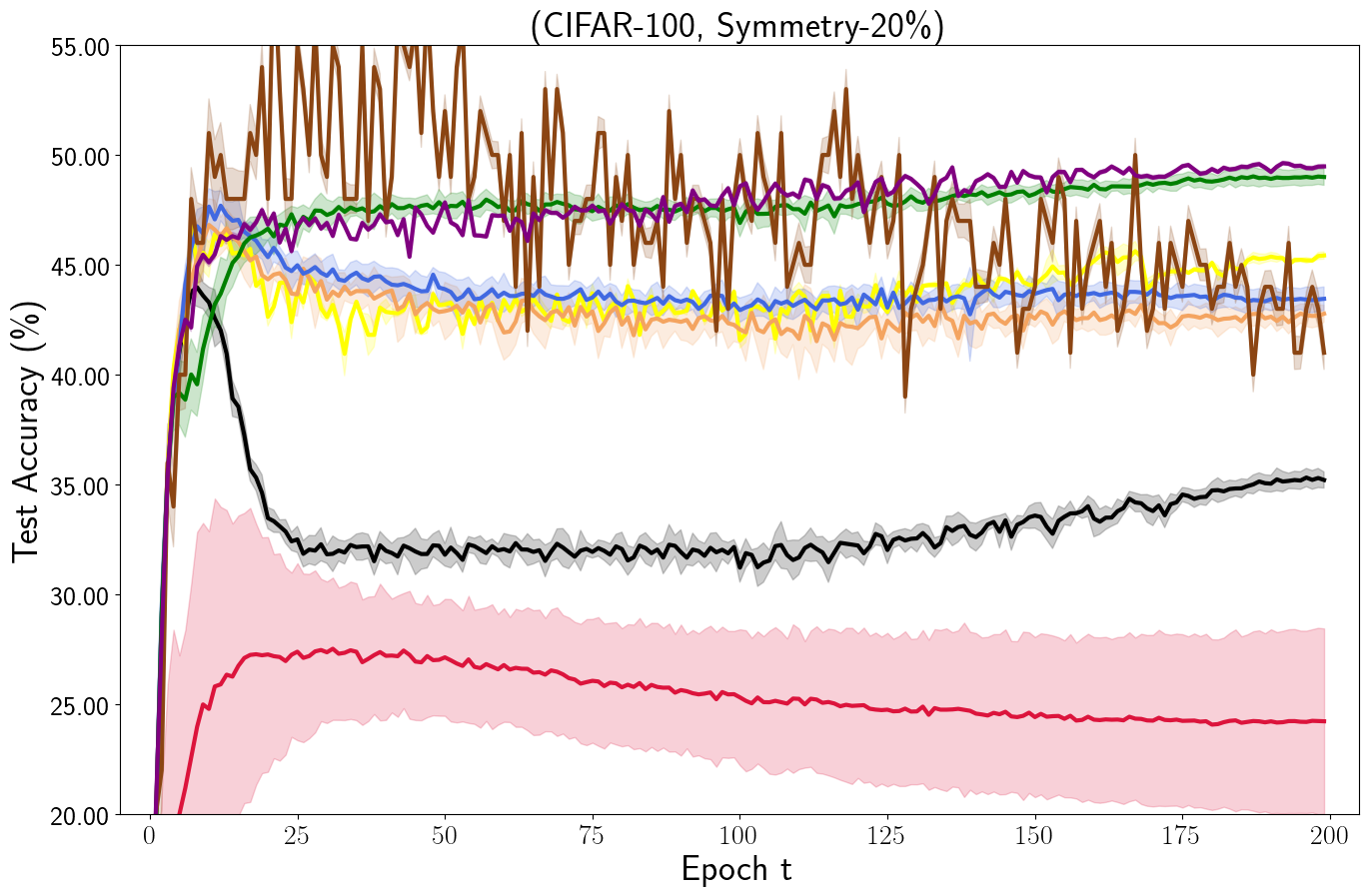}}
	\subfigure[symmetry flipping (50\%). \label{fig:dbl}]
	{\includegraphics[width=0.32\textwidth]{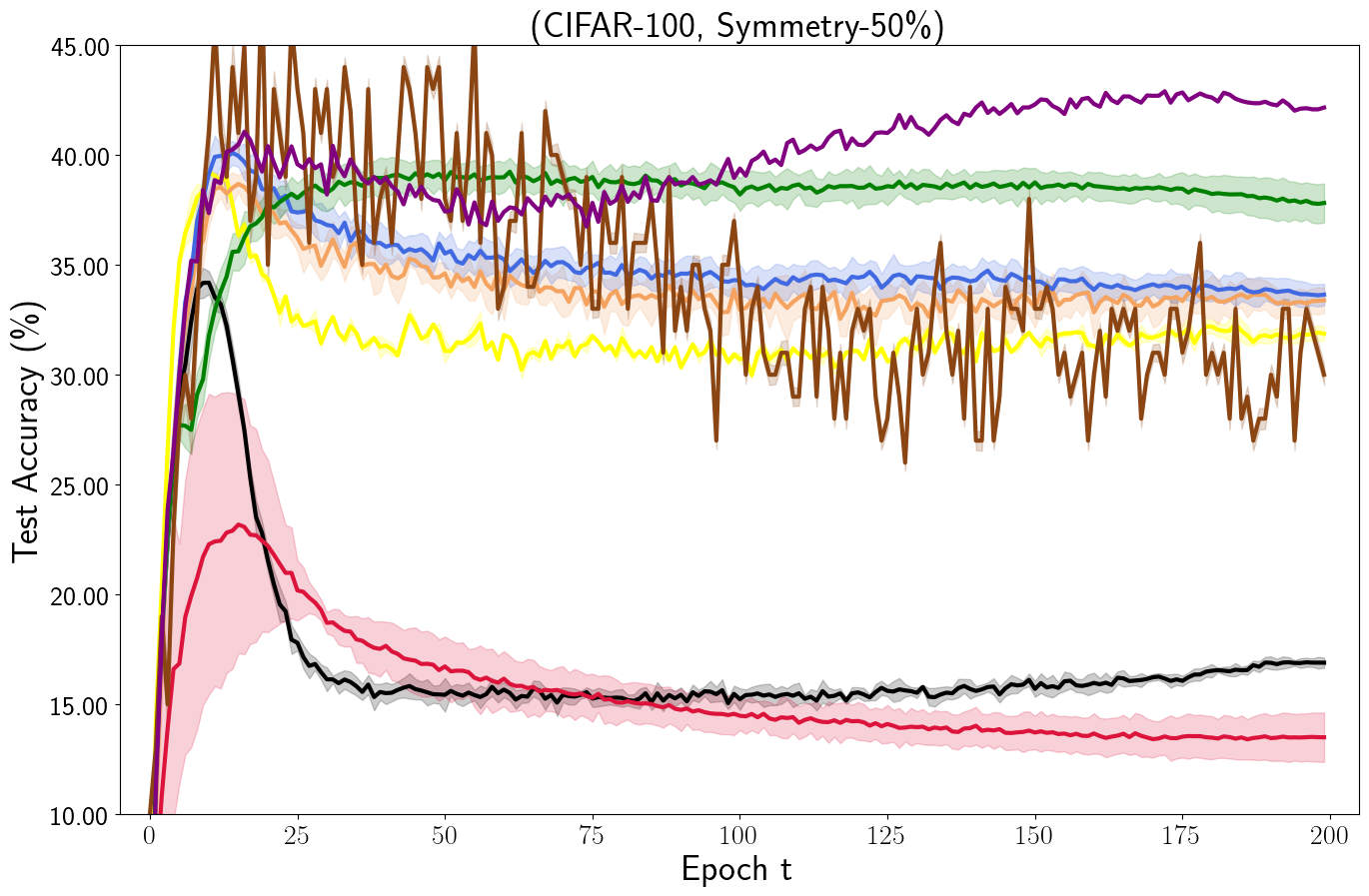}}
	\subfigure[pair flipping (45\%).]
	{\includegraphics[width=0.32\textwidth]{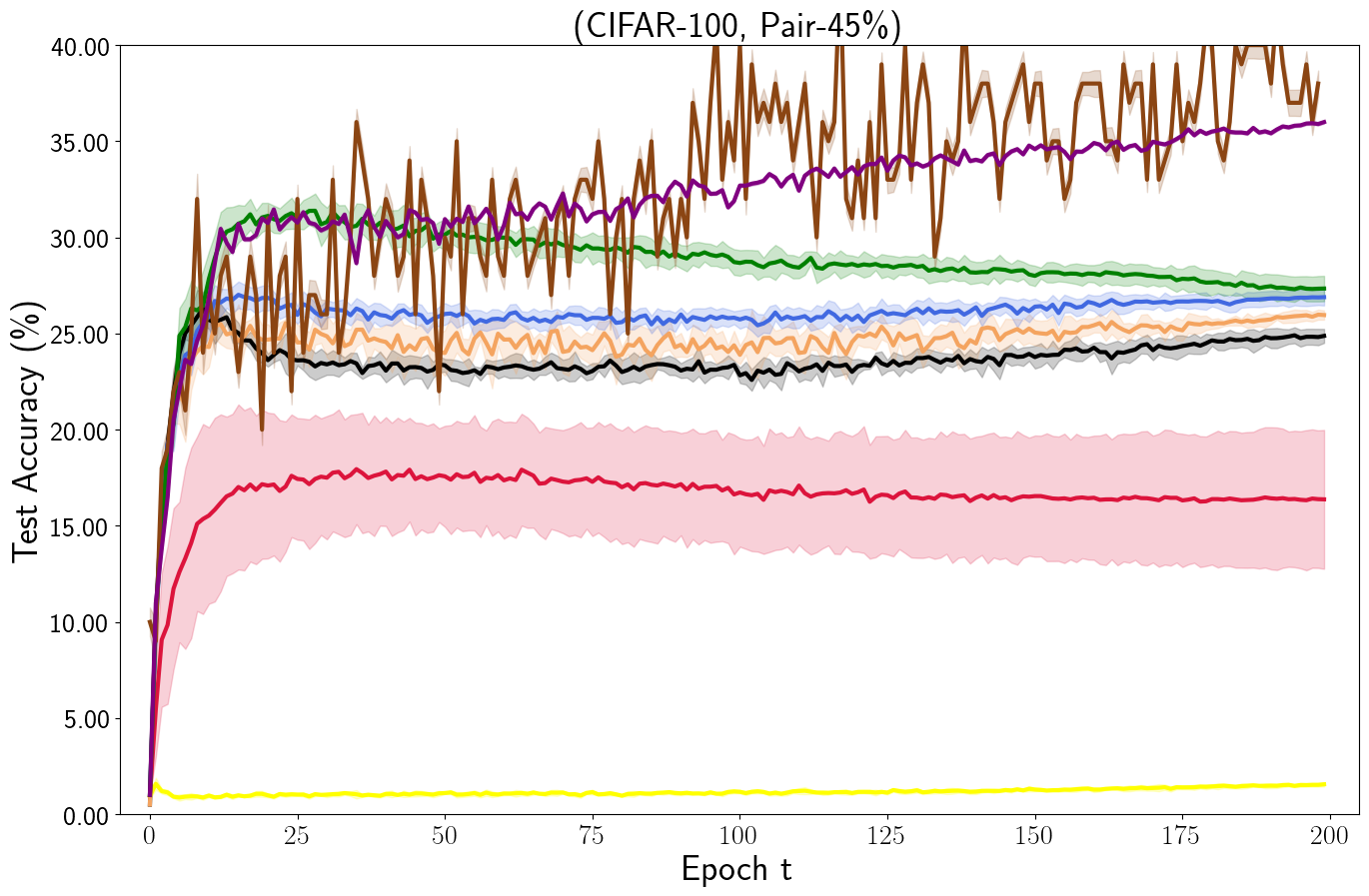}}
	
	\caption{Testing accuracies (mean and standard deviation) 
		on  MNIST (top), CIFAR-10 (middle) and CIFAR-100 (bottom).}
	\label{fig:human}
\end{figure*}

\begin{figure*}[ht]
	\centering
	
	\subfigure[symmetry flipping (20\%).]
	{\includegraphics[width=0.32\textwidth]{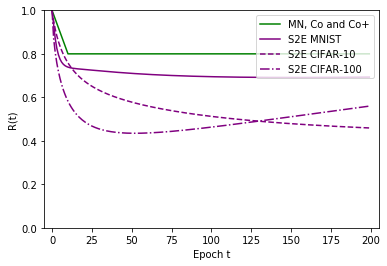}}
	\subfigure[symmetry flipping (50\%).]
	{\includegraphics[width=0.32\textwidth]{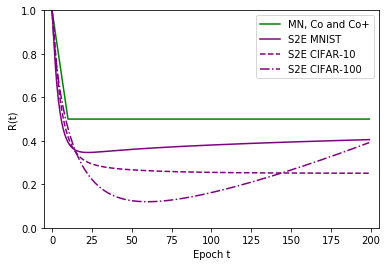}}
	\subfigure[pair flipping (45\%).]
	{\includegraphics[width=0.32\textwidth]{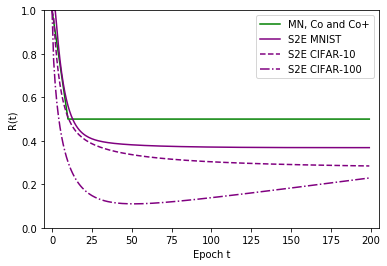}}
	
	\caption{$R(\cdot)$ obtained by the sample selection methods.  Note that \textit{MentorNet} (\textit{MN}), \textit{Co-teaching} (\textit{Co}) and \textit{Co-teaching+} (\textit{Co+})
		all use the same $R(t)$.}
	\label{fig:samrt}
\end{figure*}

\begin{figure*}[ht]
	\centering
	
	\subfigure[symmetry flipping (20\%).]
	{\includegraphics[width=0.32\textwidth]{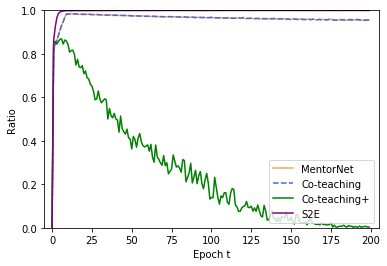}}
	\subfigure[symmetry flipping (50\%).]
	{\includegraphics[width=0.32\textwidth]{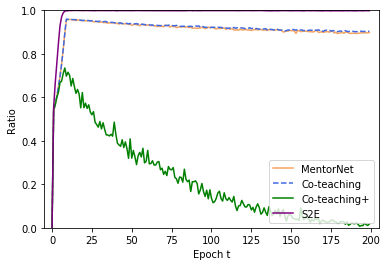}}
	\subfigure[pair flipping (45\%).]
	{\includegraphics[width=0.32\textwidth]{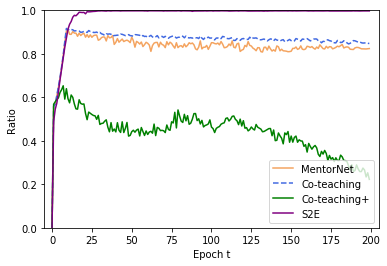}}
	
	\caption{Label precision of \textit{MentorNet}, \textit{Co-teaching}, \textit{Co-teaching+} and \textit{S2E} on MNIST.
Plots for CIFAR-10 and CIFAR-100 are in Appendix~\ref{app:otherplt}.}
	\label{fig:precision}
\end{figure*}

\begin{figure*}[ht]
	\centering
	\subfigure[symmetry flipping (20\%).]
	{\includegraphics[width=0.32\textwidth]{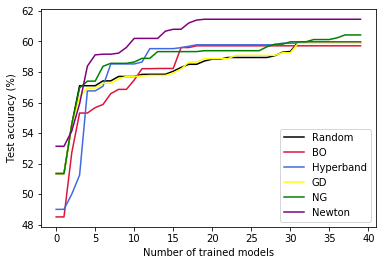}}
	\subfigure[symmetry flipping (50\%).]
	{\includegraphics[width=0.32\textwidth]{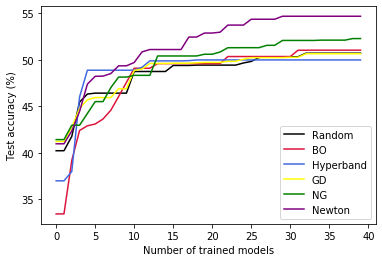}}
	\subfigure[pair flipping (45\%).]
	{\includegraphics[width=0.3275\textwidth]{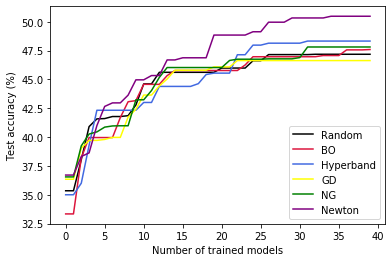}}
	
	\caption{Search efficiency of \textit{S2E} and the other search algorithms.}
	\label{fig:grad}
\end{figure*}

The following Theorem
shows that we can obtain an approximate critical
point for which the gradient norm 
is bounded by a constant factor of the gradient error. 
As $\bar{\varepsilon} = 0$
when $K \rightarrow \infty$, 
Theorem~\ref{thm:conv} ensures that a
limit point can be obtained.

\begin{theorem} \label{thm:conv}
Assume that $2 - L \eta +  \eta^2$ and
$2\eta^2 + L \eta -2$ are non-negative.
Then,
(i) For every bounded
sequence $\{ \bm{\theta}^m \}$ generated by Algorithm~\ref{alg:newton},
	there exists a limit point $\bar{\bm{\theta}}$ such that
	$\NM{\nabla \mathcal{J} (\bar{\bm{\theta}})}{} \le c_1
	\bar{\varepsilon}$,
	where $c_1$ is a positive constant.
(ii) If $\{ \bm{\theta}^m \}$ converges,
	then
	$\lim_{m \rightarrow \infty}
	\| \bm{e}^m \|
	\le 
	c_2 \bar{\varepsilon}$,
	where $c_2$ is a positive constant.
\end{theorem}
Proofs are in Appendix~\ref{app:proofs},
and are inspired by \cite{sra2012scalable,schmidt2011convergence,yao2016efficient}.
However,
they do not consider stochastic relaxation and the use of Hessian.

\section{Experiments}
\label{others}

In this section,
we demonstrate the superiority of the proposed 
{\em Search to Exploit\/} (S2E) algorithm
over the state-of-the-art in combating noisy labels. 
In step~5 of Algorithm~\ref{alg:newton}, we
use Co-teaching 
as Algorithm~\ref{alg:gen}.
Experiments are performed on standard benchmark data sets.
All the codes are  implemented in PyTorch 0.4.1, and run on a GTX 1080 Ti GPU.

\subsection{Benchmark Comparison}
\label{ssec:comp}

In this experiment, 
we use
three popular benchmark data sets:
MNIST, 
CIFAR-10 
and CIFAR-100.
Following \citep{Patrini2017,han2018co},
we add two types of label noise:
(i) symmetric flipping,
which flips the label 
to other incorrect labels
with equal probabilities;
and
(ii) pair flipping,
which flips 
a pair
of similar labels.
We use the same network architectures
as in \citep{yu2019does}.
The detailed experimental setup is in Appendix~\ref{app:exptdtl1}. 

\subsubsection{Learning Performance}

We compare
the proposed \textit{S2E} 
with the following 
state-of-the-art methods:
(i) \textit{Decoupling} \citep{malach2017decoupling};
(ii) \textit{F-correction} \citep{Patrini2017};
(iii) \textit{MentorNet} \citep{jiang2017mentornet};
(iv) \textit{Co-teaching} \citep{han2018co};
(v) \textit{Co-teaching+} \citep{yu2019does};
and 
(vi) \textit{Reweight} \citep{ren2018learning}.
As a simple baseline, we also compare with a standard deep network 
(denoted \textit{Standard})
that trains
directly on the full
noisy data set.
All experiments are repeated five times, 
and we report the averaged results.

As in \citep{Patrini2017,han2018co},
Figure~\ref{fig:human}
shows convergence of the testing accuracies.
As can be seen,
\textit{S2E} 
significantly 
outperforms the other methods
and is much more stable.

\subsubsection{The $R(\cdot)$ Learned}
\label{sec:r}

Figure~\ref{fig:samrt} compares
the $R(\cdot)$'s obtained by the proposed \textit{S2E} and the 
sample selection 
methods of
\textit{MentorNet},
\textit{Co-teaching} 
and \textit{Co-teaching+}. 
As can be seen,  the $R(\cdot)$'s learned by \textit{S2E}  are
dataset-specific,
while
the other methods always use the same
$R(\cdot)$. 
Besides, the $R(\cdot)$ learned on the noisier data 
is smaller
(e.g., compare symmetric-50\% vs symmetric-20\%). 
This is intuitive since a higher noise level means there are fewer clean
samples (smaller $R(\cdot)$) in each mini-batch.
Moreover, the proportion of large-loss samples dropped by  $R(\cdot)$ is larger
than the underlying noise level.  Intuitively, a large-loss sample usually has a
larger gradient, and can have significant impact on the model
if its label is wrong.
As 
a large-loss sample may not necessarily be noisy because the model is not perfect,
more samples are dropped.

\begin{table*}[ht]
	\centering
	\vspace{-10px}
	\caption{Best testing accuracy obtained by the various search space designs.}
	\begin{tabular}{cc | c | C{30px} | C{30px} | C{30px} | C{30px}}
		\hline
		          &   noise   &  Co-teaching   &  Single  &  RBF  &  MLP  &      S2E       \\ \hline
		          & symmetry-20\%  & 97.83 & 97.67 & 96.94 & 97.69 & \textbf{97.87} \\ \cline{2-7}
		  MNIST   & symmetry-50\%  & 96.54 & 96.56 & 95.53 & 96.16 & \textbf{96.90} \\ \cline{2-7}
		          & pair-45\% & 93.27 & 94.99 & 89.37 & 93.25 & \textbf{95.47} \\ \hline
		          & symmetry-20\%  & 57.24 & 57.83 & 56.58 & 56.82 & \textbf{58.73} \\ \cline{2-7}
		CIFAR-10  & symmetry-50\%  & 47.14 & 47.81 & 45.15 & 46.18 & \textbf{50.82} \\ \cline{2-7}
		          & pair-45\% & 44.87 & 45.19 & 42.61 & 44.26 & \textbf{47.58} \\ \hline
		          & symmetry-20\%  & 44.89 & 44.93 & 44.24 & 44.57 & \textbf{45.32} \\ \cline{2-7}
		CIFAR-100 & symmetry-50\%  & 36.53 & 36.71 & 30.99 & 35.88 & \textbf{38.74} \\ \cline{2-7}
		          & pair-45\% & 27.30 & 31.25 & 27.96 & 28.06 & \textbf{32.44} \\ \hline
	\end{tabular}
	\label{tab:space}
\end{table*}

On the other hand, 
simply dropping more samples can lead to lower accuracy
(as demonstrated in Table~8 of \citep{han2018co}).
Following \cite{han2018co},
Figure~\ref{fig:precision}
compares the label precision 
(i.e., 
ratio 
of clean samples in each mini-batch after selection)
of \textit{S2E} and other compared methods.
As can be seen,
\textit{S2E}'s label precision is consistently the highest.
This shows that the training samples used by \textit{S2E} are cleaner,
and thus yield better performance.

\subsection{Ablation Study}

\subsubsection{Search Space}
\label{sec:aba:space}

In this experiment, we study different search space designs using the
data sets in Section~\ref{ssec:comp}.
The search space of \textit{S2E} 
is compared with
(i) 
\textit{Co-teaching}:
the space specified in 
(\ref{eq:han});
and (ii) 
\textit{Single}:
the space spanned by a single basis function
in Table~\ref{tab:expfi}.
Here, we report the best performance over the four basis functions;
(iii) 
\textit{RBF}:
the space of functions output by 
a radial basis function 
network, with
one input
(epoch $t$), 
a RBF layer,
and a sigmoid output unit.  
(iv) 
\textit{MLP}:
the space of functions output by 
a multilayer perceptron 
with
one input,
a single hidden layer of ReLU units, 
and a sigmoid output unit;
The numbers of hidden units in the MLP and RBF are set to four,
which is equal to the number of basis functions in \textit{S2E}.
For a fair comparison, 
random search
is used in this experiment. This is repeated
50 times, and the average results reported.

\begin{figure}[ht]
	\centering
	\includegraphics[width=0.325\textwidth]{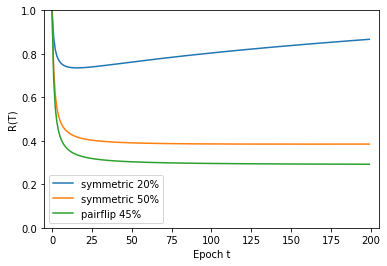}
	\vspace{-8px}
	\caption{$R(t)$ obtained by MLP on CIFAR-10.}
	\label{fig:dif:rt2}
	\vspace{-8px}
\end{figure}

Table~\ref{tab:space} shows
the best testing accuracy over all epochs obtained by the various search space variants.
\textit{Co-teaching} and \textit{Single}
perform better than 
the two general function approximators 
(\textit{RBF} and \textit{MLP}),
as their search spaces encapsulate the prior knowledge that  
$R(\cdot)$ should be of the form in Assumption~\ref{ass:prior}.
Figure~\ref{fig:dif:rt2}
shows the 
$R(\cdot)$ obtained  by
\textit{MLP} (which outperforms \textit{RBF})
on the CIFAR-10
data set (results on MNIST and CIFAR-100 are similar).
As can be seen, the shapes generally follow that in Assumption~\ref{ass:prior},
providing further empirical evidence to support
this Assumption.
The performance attained by \textit{S2E} 
is still the best
(even though only random search is used here).
This demonstrates the expressiveness and compactness of the proposed search space.

\subsubsection{Search Algorithm}
\label{sec:automl:algs}

Recall that
\textit{S2E} uses 
stochastic relaxation 
with 
Newton's method
(denoted \textit{Newton})
as the search algorithm.
In this section,
we study 
the use of other 
gradient-based search algorithms, including
(i) gradient descent (\textit{GD})
\cite{liu2018darts};  and
(ii) natural gradient descent (\textit{NG}) \cite{amari1998natural};
and also
derivative-free search algorithms, including
(i) random search (\textit{random}) \citep{bergstra2012random}; 
(ii) Bayesian optimization (\textit{BO})
\cite{bergstra2011algorithms};  and
(iii) \textit{hyperband} \citep{li2017hyperband}.
For fairness and consistency,
all these are used with Co-teaching as in  previous experiments.
We do not compare
with reinforcement learning \citep{zoph2017neural},
as our search problem does not involve a sequence of actions.
The	experiment is performed on the CIFAR-10.

In Algorithm~\ref{alg:newton}, the 
most expensive part is step~5 where Algorithm~\ref{alg:gen} is called and model training is required.
Figure~\ref{fig:grad} shows the testing accuracy w.r.t. the number of such calls.
As can be seen,
\textit{S2E}, 
with the use of the Hessian matrix,
is 
most efficient than the other algorithms compared.

\section{Conclusion}

In this paper,
we address the problem of learning with noisy labels by 
exploiting 
deep networks'
memorization effect 
with automated machine learning (AutoML).
We first design an expressive but compact 
search space
based on observations from the learning curves.
An efficient
search algorithm,
based on stochastic relaxation and Newton's method,
overcomes the difficulty of computing the gradient and allows incorporation of
information from the model and optimization objective.
Extensive experiments
on benchmark data sets
demonstrate
that
the proposed method
outperforms the state-of-the-art, and can select a higher
proportion of clean samples than other sample selection methods. 

\cleardoublepage

\section*{Acknowledgment}

This work is performed when Hansi Yang was an intern in 4Paradigm Inc supervised by Quanming Yao.
Dr. Bo Han was partially supported by the Early Career Scheme (ECS) through the Research Grants Council of Hong Kong under Grant No.22200720, HKBU Tier-1 Start-up Grant and HKBU CSD Start-up Grant.

\bibliography{bib}
\bibliographystyle{icml2020}

\end{document}